\documentclass[conference]{IEEEtran}
\IEEEoverridecommandlockouts
\usepackage{cite}
\usepackage{amsmath,amssymb,amsfonts}
\usepackage{algorithmic}
\usepackage{graphicx}
\usepackage{textcomp}
\usepackage{xcolor}
\usepackage{url}
\def\BibTeX{{\rm B\kern-.05em{\sc i\kern-.025em b}\kern-.08em
    T\kern-.1667em\lower.7ex\hbox{E}\kern-.125emX}}
\begin{document}

\title{Calibrated Alarm Thresholds for Compact Fire and Smoke Detectors}

\author{\IEEEauthorblockN{Vikas Sandireddy}
\IEEEauthorblockA{
\textit{California, USA}\\
vikas.sandireddy@jessup.edu}
}

\maketitle

\begin{abstract}
Compact object detectors such as YOLOv8n are a common choice for camera-based fire and smoke detection on low-power devices. These detectors are usually evaluated with box-level metrics such as mean average precision, and their outputs are turned into decisions with a fixed confidence threshold such as 0.25 or 0.5. A fire monitoring system, however, makes one decision per image: raise an alarm or not. In this paper we evaluate two publicly available compact detectors on the D-Fire test set at the image level. We compare three simple rules for turning box detections into one alarm score per image, and we use a split conformal procedure to choose the alarm threshold from held-out images so that the expected fraction of missed fire and smoke images stays below a chosen target. We find that the same fixed threshold leads to very different miss rates for the two detectors, 17.3 and 8.4 percent at a threshold of 0.5, while the calibrated threshold meets miss-rate targets of 10, 5 and 1 percent for both detectors and all three rules. At a 5 percent target, the false alarm rate on images without fire or smoke stays between 1.2 and 3.0 percent. We also find that smoke-only images stay below the target even when the overall target is met, and we describe the main failure cases.
\end{abstract}


\section{Introduction}

Early detection of fire and smoke can limit damage and save lives. Camera-based detection is attractive because cameras are cheap and already installed in many places. Early vision methods used hand-crafted color rules for flame pixels \cite{CELIK2009147}. Later work moved to convolutional neural networks (CNNs) \cite{8307064}, and today most systems use object detectors from the YOLO family \cite{redmon2016lookonceunifiedrealtime}. Small variants of these detectors, such as YOLOv5n or YOLOv8n, are popular because they can run on low-power devices \cite{devenancio2022dfire, fire8010026}.

Most papers in this area report box-level metrics such as precision, recall and mean average precision (mAP) on benchmarks like D-Fire \cite{devenancio2022dfire}. A deployed system, however, has to answer a simpler question for every frame: is there fire or smoke in this image or not? Two choices sit between the detector output and this answer. The first is how to turn a set of boxes into one decision. The second is which confidence threshold to use. In practice the threshold is usually fixed, for example at 0.25 \cite{feng2026hife}, and evaluations of detector calibration often fix it at 0.3 or 0.5 \cite{10.1007/978-3-031-72664-4_11}. Detector confidence scores are not calibrated probabilities \cite{3305381.3305518}, so the same threshold can mean different things for different models. For a fire alarm, the most important quantity is the fraction of fire and smoke images that are missed, and a fixed threshold gives no control over it.

This work examines these two choices for compact detectors on D-Fire and uses a split conformal procedure \cite{10.1561/2200000101} to choose the alarm threshold. The rest of the paper is organized as follows. Section~\ref{sec:related} reviews related work, Section~\ref{sec:method} describes the dataset, the detectors and the threshold selection method, Section~\ref{sec:results} presents and discusses the results, and Section~\ref{sec:concl} concludes the paper.

\section{Related Work}\label{sec:related}

\subsection{Vision-Based Fire and Smoke Detection}

Celik and Demirel \cite{CELIK2009147} proposed a rule-based color model in the YCbCr color space for classifying flame pixels and reported fire detection rates of up to 99\% on their test images. Rule-based color methods are sensitive to changes in brightness and to shadows, which can cause false alarms \cite{8307064}. Muhammad et al. \cite{8307064} proposed a CNN inspired by GoogleNet for fire detection in CCTV surveillance videos. De Ven\^{a}ncio et al. \cite{devenancio2022dfire} released the D-Fire dataset of 21,527 images, trained a YOLOv4 detector on it, and pruned its convolutional filters so that it could run on a Raspberry Pi 4. Alkhammash \cite{fire8010026} compared YOLOv9t, YOLOv10n and YOLOv11n for smoke and fire detection on two datasets, one of which is D-Fire. Feng et al. \cite{feng2026hife} proposed HIFE-YOLO, a lightweight detector based on YOLO11n, and trained it on D-Fire. Besides box-level metrics, they report a background false alarm rate, defined as the share of images without any annotation that receive at least one predicted box at a confidence of 0.25. Outside D-Fire, Dewangan et al. \cite{rs14041007} treat wildfire smoke detection from fixed cameras as a binary image classification problem and note that clouds, fog and haze are a common source of false positives.

\subsection{Post-Detection Processing and Calibration}

Joshi et al. \cite{11541411} proposed an uncertainty-aware post-detection framework for compact fire and smoke detectors. A small Confidence Refinement Network combines a dropout-based uncertainty estimate with color, edge and texture features of each box and outputs a refined confidence score. Applied to YOLOv5n and YOLOv8n on D-Fire, it gave higher precision and recall than non-maximum suppression (NMS), Soft-NMS and several color- and edge-based filters, at a modest extra processing time. The refined scores are then thresholded at a fixed value of 0.5. Our work is complementary: we leave the scores unchanged and study how the final threshold is chosen.

Guo et al. \cite{3305381.3305518} showed that modern neural networks are often poorly calibrated, meaning that their confidence scores do not match their actual accuracy. For object detectors, Kuzucu et al. \cite{10.1007/978-3-031-72664-4_11} point out that choosing an operating threshold is not trivial, that common evaluations fix it at values such as 0.3 or 0.5, and that simple post-hoc calibrators work well. In fire detection, Zhang et al. \cite{zhang2025learningcalibratereliablevisual} added a differentiable calibration-error loss to fire image classifiers and reduced their calibration error on the DFAN and EdgeFireSmoke datasets. Better-calibrated scores are useful, but they do not by themselves guarantee a given miss rate.

\subsection{Error Control With Statistical Guarantees}

Conformal prediction gives finite-sample guarantees for the output of any trained model, as long as the calibration data and the test data are exchangeable \cite{10.1561/2200000101}. Conformal risk control extends this idea to the expected value of monotone losses such as the false negative rate \cite{angelopoulos2025conformalriskcontrol}. In binary classification, Tong et al. \cite{tong2018np} proposed the Neyman-Pearson umbrella algorithm, which picks the threshold from an order statistic of held-out scores of one class so that the error on that class stays below a chosen level with high probability. Luo et al. \cite{10.1177/02783649231221580} used conformal prediction to tune warning systems in robotics so that they have a provable false negative rate. For object detection, Andéol et al. \cite{andeol2025conformal} proposed sequential conformal risk control, which gives guarantees on the boxes and classes produced by a detector. In the fire domain, Dayan \cite{dayan2026conformalriskcontrolsafetycritical} applied conformal risk control to wildfire spread prediction and controlled the pixel-level false negative rate. Our setting is different from these works: we control the miss rate of an image-level alarm built on top of a camera-based fire and smoke detector.

\begin{figure}[t]
\centerline{\includegraphics[width=\columnwidth]{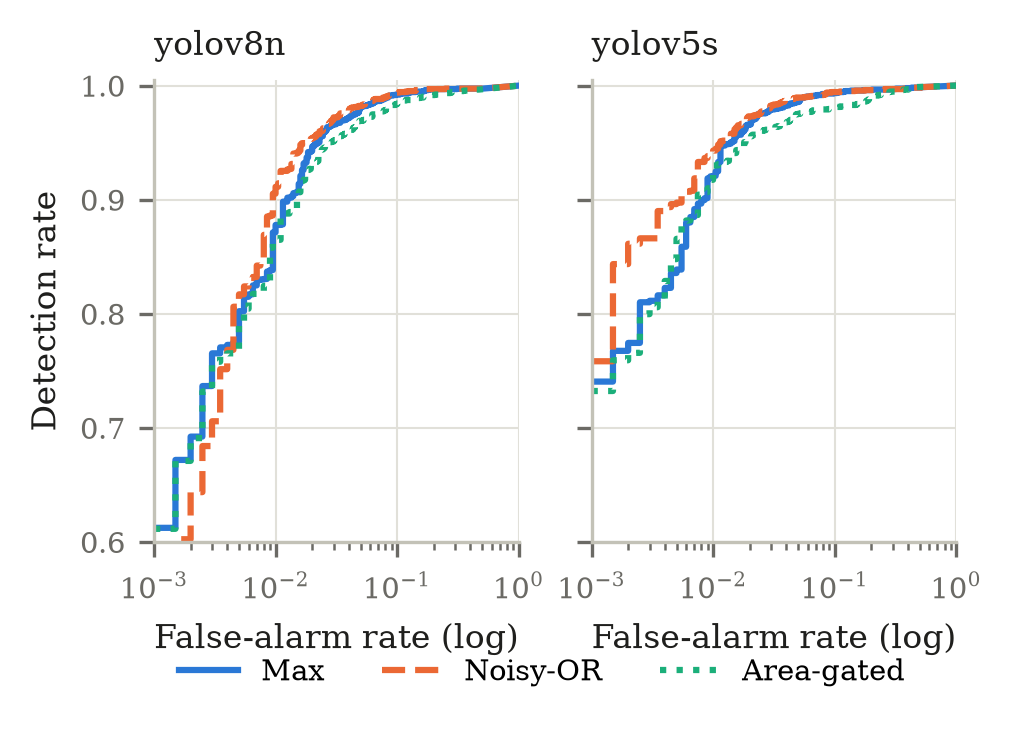}}
\caption{Image-level ROC curves for the three rules. The horizontal axis is on a log scale to show the region of low false alarm rates.}
\label{fig:roc}
\end{figure}

\begin{figure*}[t]
\centerline{\includegraphics[width=0.85\textwidth]{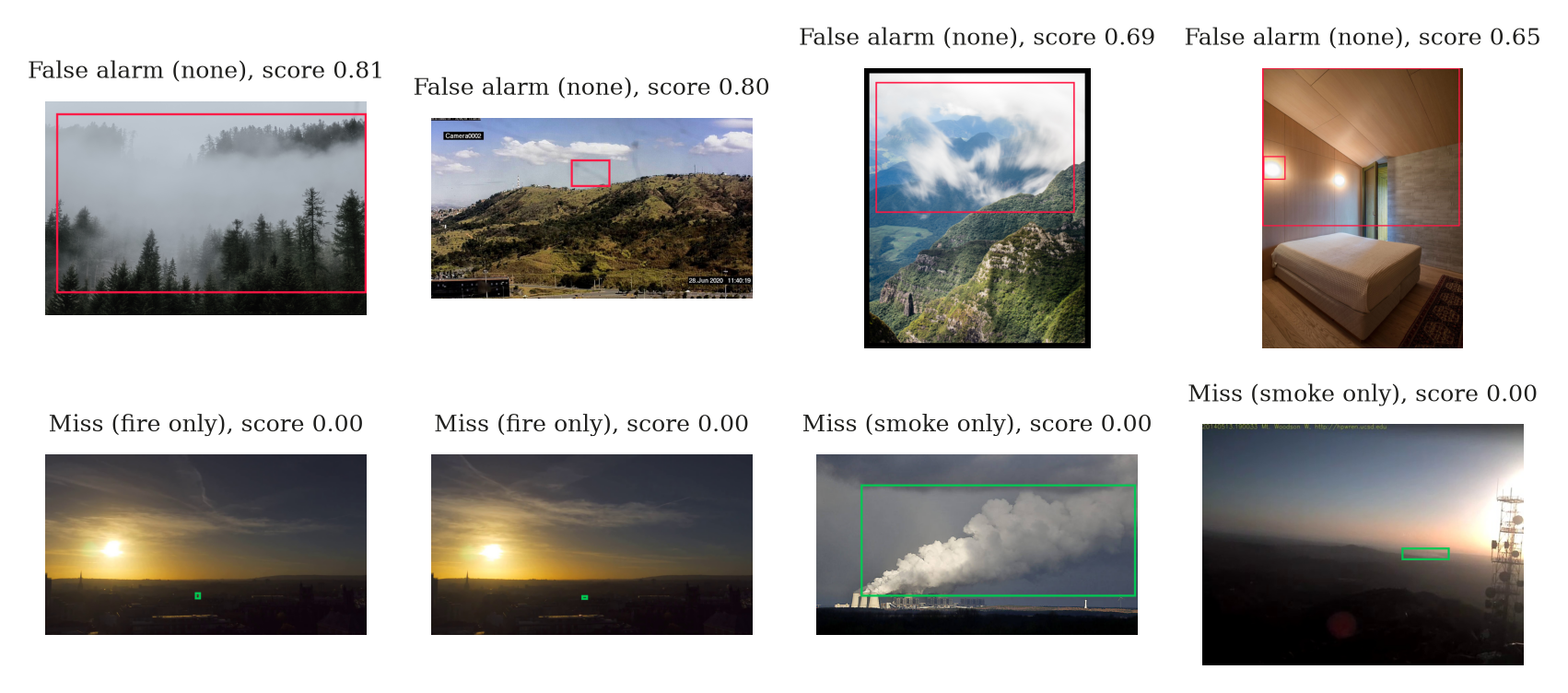}}
\caption{Failure cases of YOLOv8n with the max rule at the 5\% miss target. Top: negative images with the highest scores (false alarms). Bottom: positive images with the lowest scores (misses). Green boxes are ground truth and red boxes are predictions above the threshold.}
\label{fig:fail}
\end{figure*}

\section{Methodology}\label{sec:method}

\subsection{Dataset}

We use the D-Fire dataset \cite{devenancio2022dfire}. It contains 21,527 images in four categories: fire only, smoke only, fire and smoke, and none. Images in the last category contain neither fire nor smoke, but often show scenes that can be mistaken for them. The authors split the dataset into a training set of 17,221 images and a test set of 4,306 images. All our experiments use only the 4,306 test images, which contain 2,878 fire boxes and 2,315 smoke boxes. We call an image \emph{positive} if it has at least one fire or smoke box, and \emph{negative} otherwise.

\subsection{Detectors}

We did not train any detector. Instead we evaluate two publicly available compact detectors that were trained on D-Fire:
\begin{itemize}
\item \textbf{YOLOv8n} \cite{yolov8}, fine-tuned on D-Fire and released on Hugging Face (\url{https://huggingface.co/rabahdev/fire-smoke-yolov8n}). Its model card reports an mAP@50 of 0.754 on the D-Fire test split.
\item \textbf{YOLOv5s} \cite{yolov5}, released by the authors of D-Fire (\url{https://github.com/pedbrgs/Fire-Detection}).
\end{itemize}
Both models were run once over the test images at an input size of 640 pixels, keeping every box with a confidence of at least 0.001 and at most 100 boxes per image. For NMS we used the default IoU threshold of each framework, 0.7 for YOLOv8n and 0.6 for YOLOv5s. All later analysis works on these saved boxes, so no further inference is needed.

\subsection{Box-Level Metrics}

To place our detectors next to published results, we first compute the usual box-level metrics: precision and recall at a confidence of 0.25 with an IoU of 0.5, average precision at IoU 0.5 for each class (AP@50), and its mean over the two classes (mAP@50), as well as mAP averaged over IoU thresholds from 0.5 to 0.95 (mAP@50--95). We use greedy matching and 101-point interpolation.

\subsection{From Boxes to an Image Score}

For an image $I$, let $c_1,\dots,c_m$ be the confidences of its predicted boxes and $a_1,\dots,a_m$ their areas as a fraction of the image area. We compare three rules that map these boxes to one score $s(I)$:
\begin{align}
s_{\max}(I) &= \max_{j} c_j, \label{eq:max}\\
s_{\mathrm{nor}}(I) &= 1-\prod_{j \in T_3(I)} (1-c_j), \label{eq:nor}\\
s_{\mathrm{area}}(I) &= \max_{j:\, a_j \geq 0.001} c_j, \label{eq:area}
\end{align}
where $T_3(I)$ holds the three most confident boxes. The score is 0 when no box qualifies. Rule \eqref{eq:max} is the standard choice: an alarm is raised if any single box is confident enough. Rule \eqref{eq:nor} is the noisy-OR used in multiple instance learning \cite{NIPS2005_56c82ccd}, which lets several weak boxes add up. Rule \eqref{eq:area} ignores very small boxes, covering less than 0.1\% of the image, on the idea that they are often noise. An alarm is raised when $s(I) \geq t$ for a threshold $t$. Both classes count: an alarm is correct on a positive image whether it comes from a fire box or a smoke box.

\subsection{Image-Level Metrics}

Let $P$ and $N$ be the sets of positive and negative images. For a threshold $t$, the detection rate is $\mathrm{DR}(t) = \frac{1}{|P|}\sum_{I \in P} \mathbf{1}[s(I) \geq t]$ and the false alarm rate is $\mathrm{FAR}(t) = \frac{1}{|N|}\sum_{I \in N} \mathbf{1}[s(I) \geq t]$. The miss rate is $1-\mathrm{DR}(t)$. We also report the area under the ROC curve (AUC) of each score.

\subsection{Calibrated Threshold Selection}

We want a threshold that keeps the miss rate below a target $\alpha$, for example 5\%. Let $s_{(1)} \leq \dots \leq s_{(n)}$ be the sorted scores of $n$ positive calibration images. We set
\begin{equation}
\hat{t} = s_{(k)}, \qquad k = \lfloor \alpha (n+1) \rfloor, \label{eq:thr}
\end{equation}
and raise an alarm for every image whose score is at least $\hat{t}$ (if $k=0$, every image raises an alarm). If a new positive image is exchangeable with the calibration images, its score is equally likely to fall in any position among the $n+1$ scores. It falls below $s_{(k)}$ with probability at most $k/(n+1) \leq \alpha$, so
\begin{equation}
\Pr\big[s(I_{\mathrm{new}}) < \hat{t}\,\big] \leq \alpha.
\end{equation}
This is the split conformal argument \cite{10.1561/2200000101} applied to positive images only. The guarantee holds on average over the choice of calibration set. The Neyman-Pearson umbrella algorithm \cite{tong2018np} uses the same kind of order statistic but picks $k$ so that the guarantee holds with high probability, which needs more data. Only positive images are needed to set $\hat{t}$, and the false alarm rate is whatever the detector delivers at that threshold.

Since we have only one labeled split, we repeat the following 200 times. We randomly divide the 4,306 test images into two halves, keeping the share of each category the same in both. We compute $\hat{t}$ from the positive images in the first half and measure DR and FAR on the second half. We report the mean and standard deviation over the 200 repeats for $\alpha \in \{0.10, 0.05, 0.01\}$, and compare with the fixed thresholds 0.25 and 0.5 on the same halves.

\section{Results and Discussion}\label{sec:results}

\subsection{Box-Level Results}

Table~\ref{tab:box} shows the box-level metrics. Our mAP@50 for YOLOv8n is 0.746, close to the 0.754 reported on its model card, which suggests that our evaluation matches the standard one. For both detectors, fire is the harder class: its AP@50 is 0.684 and 0.720, compared with 0.807 and 0.822 for smoke. Joshi et al. \cite{11541411} observe the same pattern for compact YOLOv5n and YOLOv8n detectors on D-Fire, and for YOLOv5n their post-detection framework gave its largest improvement on the fire class. This suggests that fire remains the weaker class for compact detectors on this dataset.

\begin{table}[t]
\caption{Box-Level Results on the D-Fire Test Set}
\begin{center}
\begin{tabular}{|l|c|c|}
\hline
\textbf{Metric} & \textbf{YOLOv8n} & \textbf{YOLOv5s} \\
\hline
Precision & 0.685 & 0.678 \\
Recall & 0.724 & 0.781 \\
mAP@50 & 0.746 & 0.771 \\
mAP@50--95 & 0.424 & 0.429 \\
AP@50, fire & 0.684 & 0.720 \\
AP@50, smoke & 0.807 & 0.822 \\
\hline
\end{tabular}
\label{tab:box}
\end{center}
\end{table}

\subsection{Image-Level Results With Fixed Thresholds}

Table~\ref{tab:fixed} shows the image-level results. Even though mAP@50 is only around 0.75, the AUC at the image level is between 0.990 and 0.996. Many errors that hurt box-level metrics, such as a box that is slightly misplaced or a missed second box in the same image, do not matter for an alarm.

The fixed thresholds, however, behave very differently for the two detectors. With the max rule at $t=0.5$, YOLOv8n misses 17.3\% of positive images while YOLOv5s misses 8.4\%. At $t=0.25$ the miss rates are 4.3\% and 2.3\%. The same threshold thus gives a twofold difference in missed fires, and without labeled data the user cannot know this. At $t=0.25$, the false alarm rate of the max rule is 2.4\% for YOLOv8n and 2.9\% for YOLOv5s. This is the same quantity that Feng et al. \cite{feng2026hife} call the background false alarm rate, and their value for YOLO11n on D-Fire, 3.24\%, is in the same range. The noisy-OR rule raises scores, so at a fixed threshold it detects more but also raises more false alarms, which again shows that a fixed threshold has no fixed meaning.

\begin{table}[t]
\caption{Image-Level Results With Fixed Thresholds}
\begin{center}
\resizebox{\columnwidth}{!}{%
\begin{tabular}{|l|l|c|c|c|c|c|}
\hline
\textbf{Model} & \textbf{Rule} & \textbf{AUC} & \multicolumn{2}{c|}{$\boldsymbol{t=0.25}$} & \multicolumn{2}{c|}{$\boldsymbol{t=0.5}$} \\
\cline{4-7}
 & & & \textbf{DR} & \textbf{FAR} & \textbf{DR} & \textbf{FAR} \\
\hline
YOLOv8n & Max & 0.993 & 0.957 & 0.024 & 0.827 & 0.007 \\
 & Noisy-OR & 0.994 & 0.972 & 0.032 & 0.924 & 0.011 \\
 & Area-gated & 0.990 & 0.942 & 0.023 & 0.815 & 0.006 \\
\hline
YOLOv5s & Max & 0.995 & 0.977 & 0.029 & 0.916 & 0.009 \\
 & Noisy-OR & 0.996 & 0.987 & 0.038 & 0.958 & 0.014 \\
 & Area-gated & 0.992 & 0.959 & 0.025 & 0.901 & 0.007 \\
\hline
\end{tabular}}
\label{tab:fixed}
\end{center}
\end{table}

\subsection{Calibrated Thresholds}

Table~\ref{tab:conf} shows the results of the calibrated thresholds over the 200 random splits. In all 18 settings (two detectors, three rules, three targets), the mean miss rate on the evaluation halves is at or just below the target $\alpha$. The detection rate varies from split to split with a standard deviation of about 0.01, so a single split can land slightly above the target. This is expected, since the guarantee holds on average over calibration sets \cite{10.1561/2200000101}.

The thresholds themselves show why calibration is needed. For a 5\% miss target with the max rule, the median calibrated threshold is 0.277 for YOLOv8n and 0.420 for YOLOv5s. A fixed threshold of 0.5 is too high for both, and a single fixed value cannot suit both models.

The cost in false alarms is small for moderate targets and grows quickly for strict ones. At a 5\% target, the false alarm rate is between 1.2\% and 3.0\%. At a 1\% target, it rises to between 6.7\% and 21.9\%. Among the rules, noisy-OR gives the lowest false alarm rate at the 5\% target for both detectors, as Fig.~\ref{fig:roc} also suggests at low false alarm rates. The area-gated rule is the worst at the 1\% target (19.1\% and 21.9\% false alarms). Removing small boxes removes real evidence: some of the hardest positive images contain only a small, distant fire or a thin smoke plume, and the threshold must then drop to almost zero to catch them.

\begin{table}[t]
\caption{Results With Calibrated Thresholds}
\begin{center}
\resizebox{\columnwidth}{!}{%
\begin{tabular}{|l|l|c|c|c|c|c|c|}
\hline
\textbf{Model} & \textbf{Rule} & \multicolumn{2}{c|}{$\boldsymbol{\alpha=0.10}$} & \multicolumn{2}{c|}{$\boldsymbol{\alpha=0.05}$} & \multicolumn{2}{c|}{$\boldsymbol{\alpha=0.01}$} \\
\cline{3-8}
 & & \textbf{Miss} & \textbf{FAR} & \textbf{Miss} & \textbf{FAR} & \textbf{Miss} & \textbf{FAR} \\
\hline
YOLOv8n & Max & 0.098 & 0.013 & 0.048 & 0.023 & 0.009 & 0.092 \\
 & Noisy-OR & 0.099 & 0.010 & 0.048 & 0.019 & 0.009 & 0.084 \\
 & Area-gated & 0.099 & 0.015 & 0.049 & 0.030 & 0.009 & 0.191 \\
\hline
YOLOv5s & Max & 0.099 & 0.008 & 0.049 & 0.014 & 0.009 & 0.069 \\
 & Noisy-OR & 0.099 & 0.005 & 0.049 & 0.012 & 0.009 & 0.067 \\
 & Area-gated & 0.099 & 0.008 & 0.049 & 0.019 & 0.010 & 0.219 \\
\hline
\end{tabular}}
\label{tab:conf}
\end{center}
\end{table}

\subsection{Results per Category}\label{sec:cat}

\begin{figure}[t]
\centerline{\includegraphics[width=\columnwidth]{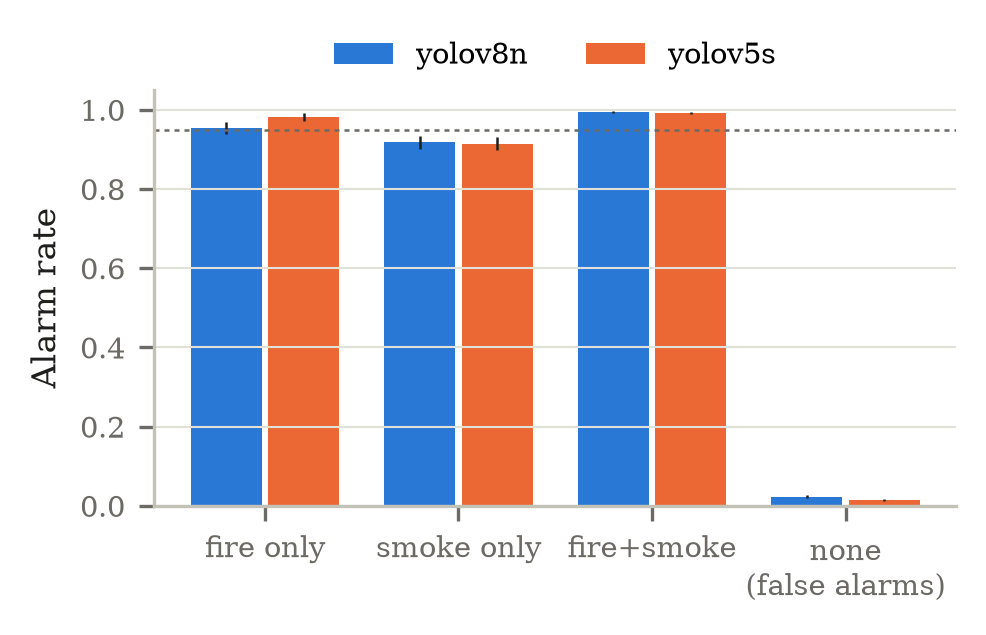}}
\caption{Alarm rate per D-Fire category with the max rule at a 5\% miss target (mean and standard deviation over 200 splits). For the first three categories this is the detection rate; for ``none'' it is the false alarm rate. The dotted line marks 0.95.}
\label{fig:cat}
\end{figure}

The guarantee covers positive images as a whole, not each type of image. Fig.~\ref{fig:cat} breaks the alarm rate down by D-Fire category at the 5\% target with the max rule. Images with both fire and smoke are almost always detected (0.996 for YOLOv8n and 0.992 for YOLOv5s), and fire-only images are above the target (0.954 and 0.981). Smoke-only images, however, are detected only 91.8\% and 91.4\% of the time, below the 95\% target. The overall target is met because the easier categories make up for them. For a system whose main purpose is early warning, where faint smoke is often the first sign of a fire, this matters.

\subsection{Failure Cases}

Fig.~\ref{fig:fail} shows the four negative images with the highest scores and the four positive images with the lowest scores for YOLOv8n. The false alarms are mostly things that look like smoke or fire: a forest in fog, clouds over mountains, and a warm wall lamp in a bedroom. This matches the observation of Dewangan et al. \cite{rs14041007} that clouds, fog and haze cause false positives. The highest-scoring negative images of YOLOv5s (not shown) are similar and also include street lights at night and decorative lights. Several false alarms on both detectors come from fixed hillside cameras, where the detector marks a small spot near the horizon. We cannot tell from the images whether some of these are distant smoke that was not labeled.

The misses fall into two groups. Some are very small targets, such as a tiny fire on the horizon at sunset or a faint plume in haze. Others have labels that are arguably ambiguous, such as a large white plume from the cooling towers of a power plant that is labeled as smoke. All of the misses shown have a score of 0.00, meaning that the detector produced no box for them at all, even at a confidence of 0.001. No choice of threshold can recover such images, which explains why very strict targets lead to many false alarms. We also noticed that two of the missed images are near-duplicates of each other. Feng et al. \cite{feng2026hife} checked D-Fire for exact duplicates and found none, but did not check for near-duplicates.

\subsection{Limitations}

This study uses a single dataset and two detectors, so the results may not carry over to other datasets or models. The calibration assumes that new images resemble the calibration images, which may not hold when cameras, scenes or weather change after deployment. The guarantee also holds on average over all positive images, not for each category, as Section~\ref{sec:cat} shows for smoke-only images. Finally, we work with single images, while real systems process video and could use information over time.

\section{Conclusion}\label{sec:concl}

We studied compact fire and smoke detectors as image-level alarms. On D-Fire, two public detectors separate positive and negative images very well (AUC above 0.99), but common fixed thresholds lead to very different miss rates for different models. A simple split conformal threshold, computed from the positive images in half of the test set, met miss-rate targets of 10\%, 5\% and 1\% for both detectors and all three rules, at a false alarm rate of 1.2\% to 3.0\% for a 5\% target. Noisy-OR was the best of the three rules, and ignoring small boxes hurt at strict targets. Smoke-only images remained below the target. In future work, we plan to calibrate separate thresholds per category, combine calibration with post-detection rescoring methods, test the approach on other datasets and on video, and study high-probability guarantees such as the Neyman-Pearson approach \cite{tong2018np}.

\bibliographystyle{IEEEtran}
\bibliography{refs}

\end{document}